\documentclass[twoside,11pt]{article}

\usepackage{jmlr2e}
\usepackage{color}
\usepackage{listings}
\usepackage[utf8]{inputenc}
\usepackage{wrapfig}
\usepackage{amssymb}

\usepackage{url}
\usepackage{todonotes}
\usepackage{xspace}
\usepackage{array}

\usepackage{microtype}

\newcommand{\kymatio}{\textit{Kymatio}\xspace}

\newcommand\Edouard[1]{}

\newcommand{\conv}{\circledast}
\newcommand{\wavu}{\psi^{(1)}}
\newcommand{\wavd}{\psi^{(2)}}
\newcommand{\low}{\phi}
\newcommand{\la}{{\lambda}}
\newcommand{\lau}{{\lambda_1}}
\newcommand{\lad}{{\lambda_2}}
\newcommand{\Lau}{{\Lambda_1}}
\newcommand{\Lad}{{\Lambda_2}}

\definecolor{dkgreen}{rgb}{0, 0.5, 0.1}

\usepackage{lastpage}
\jmlrheading{21}{2020}{1-\pageref{LastPage}}{1/19; Revised
10/19}{1/20}{19-047}{Mathieu Andreux, Tomás Angles, Georgios Exarchakis, Roberto Leonarduzzi, Gaspar Rochette, Louis Thiry, John Zarka, Stéphane Mallat, Joakim Andén, Eugene Belilovsky, Joan Bruna, Vincent Lostanlen, Muawiz Chaudhary, Matthew J. Hirn, Edouard Oyallon, Sixin Zhang, Carmine Cella, Michael Eickenberg}
\ShortHeadings{Kymatio: Scattering Transforms in Python}{Andreux et al.}

\begin{document}

\title{Kymatio: Scattering Transforms in Python}

\author{\name Mathieu Andreux
        \email{mathieu.andreux@owkin.com} \\
        \name Tomás Angles
        \email{tomas.angles@ens.fr} \\
        \name Georgios Exarchakis
        \email{georgios.exarchakis@ens.fr} \\
        \name Roberto Leonarduzzi
        \email{roberto.leonarduzzi@ens-lyon.fr} \\
        \name Gaspar Rochette
        \email{gaspar.rochette@ens.fr} \\
        \name Louis Thiry
        \email{louis.thiry@outlook.fr} \\
        \name John Zarka
        \email{johnzarka@gmail.com} \\
        \addr École normale supérieure, CNRS, PSL Research University, 45, rue d'Ulm, 75005 Paris, France
        \AND
        \name St\'ephane Mallat
        \email stephane.mallat@ens.fr \\
        \addr École normale supérieure, CNRS, PSL Research University, 45, rue d'Ulm, 75005 Paris, France \\
        \addr Collège de France, 11, place Marcelin-Berthelot
75231 Paris, France \\
        \addr Flatiron Institute, 162 5th Avenue, New York, NY 10010, USA
        \AND
        \name Joakim And\'en
        \email janden@flatironinstitute.org \\
        \addr Flatiron Institute, 162 5th Avenue, New York, NY 10010, USA
        \AND
        \name Eugene Belilovsky
        \email eugene.belilovsky@umontreal.ca \\
        \addr Mila, Université de Montréal, 6666 St Urbain Street, Montreal, Quebec H2S 3H1, Canada
        \AND
        \name Joan Bruna
        \email bruna@cims.nyu.edu \\
        \name Vincent Lostanlen
        \email vl1019@nyu.edu \\
        \addr New York University, 70 Washington Square South,
        New York, NY 10012, USA
        \AND
        \name Muawiz Chaudhary
        \email chaudhm@wwu.edu \\ 
        \addr Western Washington University, 516 High Street, Bellingham, WA 98225, USA
        \AND
        \name Matthew J. Hirn
        \email mhirn@msu.edu \\
        \addr Michigan State University, 426 Auditorium Road
East Lansing, MI 48824, USA
        \AND
        \name Edouard Oyallon
        \email edouard.oyallon@lip6.fr \\ 
        \addr CNRS, LIP6, Sorbonne University, 4 place Jussieu, 75252 Paris, France
        \AND
        \name Sixin Zhang
        \email sixin.zhang@pku.edu.cn \\
        \addr Peking University, No. 5 Yiheyuan Road, Haidian District, Beijing 100871, China
        \AND
        \name Carmine Cella
        \email carmine.cella@berkeley.edu \\ 
        \addr University of California, Berkeley, 101 Sproul Hall, Berkeley, CA 94720, USA
        \AND
        \name Michael Eickenberg
        \email meickenberg@flatironinstitute.org \\
        \addr Flatiron Institute, 162 5th Avenue, New York, NY 10010, USA
}

\editor{Balazs Kegl}

\maketitle

\begin{abstract}
The wavelet scattering transform is an invariant and stable signal representation suitable for many signal processing and machine learning applications. We present the \kymatio software package, an easy-to-use, high-performance Python implementation of the scattering transform in 1D, 2D, and 3D that is compatible with modern deep learning frameworks, including \textit{PyTorch} and \textit{TensorFlow}/\textit{Keras}. The transforms are implemented on both CPUs and GPUs, the latter offering a significant speedup over the former. The package also has a small memory footprint.
Source code, documentation, and examples are available under a BSD license at \url{https://www.kymat.io}.
\end{abstract}


\section{Introduction}

Many classification and regression tasks have a degree of invariance to translations and deformations, such as those relating to images, audio recordings, and electronic densities. The scattering transform was introduced in \citet{cpam} to build a signal representation that is invariant to such transformations while preserving as much as possible the information relevant to the task at hand. It is defined as a convolutional network whose filters are fixed to be wavelet and lowpass averaging filters coupled with modulus nonlinearities. It has many favorable theoretical properties \citep{cpam,bruna2015,irene} and enjoys considerable success as a powerful tool in modern signal processing \citep{adel20173d,bruna2013invariant,dss,chudacek,Sifre_2013_CVPR,solidharmonic}. It is also effective in combination with modern representation learning approaches \citep{oyallon2018scattering,sainath-dss,zeghidour}.




This article presents \kymatio, a scattering transform implementation that is user-friendly, well-documented, fast, and compatible with existing automatic differentiation libraries. It brings together transforms in 1D, 2D, and 3D under a \textit{unified} application programming interface (API). 
The scattering network is also traversed depth-first to reduce memory requirements, enabling efficient processing in limited-memory environments, such as GPUs.

\section{Implementing the Scattering Transform}


\paragraph{Definition}We consider signals defined on a grid of size $N_1 \times \cdots \times N_d$ for $d = 1, 2, 3$. Given two signals $x[n]$ and $y[n]$ on this grid, we denote their periodic convolution by $x \conv y[n]$. The second-order scattering transform is defined using two wavelet filter banks $\{\wavu_\lau[n]\}_{\lau \in \Lau}$ and $\{\wavd_\lad[n]\}_{\lad \in \Lad}$, where $\lau$ and $\lad$ are frequency indices in the sets $\Lau$ and $\Lad$. It also includes a lowpass filter $\low_J[n]$, where the integer $J > 0$ specifies the averaging scale $2^J$ of the filter. Together with a non-linearity $\rho(t)$, these filters define the scattering transform.

The zeroth-order scattering coefficient $S_0 x[n]$ is the local average given by $S_0 x[n] = x \conv \low_J[n]$.
Convolving $x[n]$ with the first-order wavelet filter bank $\{\wavu_\lau[n]\}_{\lau \in \Lau}$, applying $\rho(t)$, and convolving with $\low_J[n]$, we obtain the first-order scattering coefficients
\begin{equation}
    \nonumber
    S_1 x[n, \lau] = \rho\left(x \conv \wavu_\lau\right) \conv \low_J[n], \qquad \lau \in \Lau.
\end{equation}
The modulus of the first wavelet transform acts as a demodulation, shifting its energy to the low frequencies. However, only some of these frequencies are covered by the low-pass filter $\low_J$. 
We recover the remaining frequencies by decomposing $\rho(x \conv \wavu_\lau[n])$ using the second filter bank, but this is done only for a subset $\Lad(\lau)$ of $\Lad$ since $\rho(x \conv \wavu_\lau[n])$ is a low-frequency signal. Typically, we have $\Lad(\lau)=\{\la\in\Lad,|\la|>|\lau|\}$. The result is then passed through $\rho(t)$ and averaged, yielding the second-order coefficients
\begin{equation}
    \nonumber
    S_2 x[n, \lau, \lad] = \rho\left(\rho\left(x \conv \wavu_\lau\right) \conv \wavd_\lad\right) \conv \low_J[n], \qquad \lau \in \Lau, \lad \in \Lad(\lau).
\end{equation}

The energy of higher-order scattering coefficients is typically small and does not greatly influence results \citep{irene,bruna2013invariant,dss}. On the other hand they can be computationally intensive. We have thus chosen to restrict our scope to second-order coefficients, which is what is used in most works.

\paragraph{Implementation}
Signals obtained by filtering and applying $\rho(t)$ are low-frequency, so intermediate results are downsampled to reduce computational load as in \citet{scatnet}. In 1D and 2D, we use Morlet wavelets which are close to analytic (\textit{i.e.}, complex-valued with low energy in the negative frequencies) and the non-linearity is the complex modulus $\rho(t) = |t|$ for $t \in \mathbb{C}$ \citep{dss,bruna2013invariant}. The 3D transform is calculated using solid harmonic wavelets $\psi_{\lambda_1} = \psi_{j, \ell, m}$, where $j$ indexes the scale, and $\ell, m$ are the azimuthal and magnetic quantum numbers. In this case the non-linearity $\rho : \mathbb{C}^{2\ell+1} \rightarrow \mathbb{R}$ is defined, with a slight abuse of notation, as $\rho(x\circledast \psi_{j,\ell}) = \sqrt{\sum_m{|x\circledast \psi_{j,\ell,m}|^2}}$ \citep{solidharmonic}. Following \citet{oyallon2018scattering}, we compute the scattering transform in a depth-first manner, reducing the number of intermediate signals stored at a given time. Since convolutions are all periodic, they may be efficiently calculated using fast Fourier transforms.


\section{Project vision}

\paragraph{Code quality} Adopting the philosophy of \textit{scikit-learn} \citep{pedregosa2011scikit}, the goal of \kymatio is not to maximize the number of features, but to provide a stable and easy-to-use framework. To this end, we make heavy use of unit tests, minimize the number of dependencies, and strive for intuitive interfaces inspired by modern deep learning paradigms. 
\kymatio also provides an extensive user guide, including an API reference, a tutorial, installation instructions, and easy-to-understand examples, several of which feature real-world applications.

\paragraph{Community and bug tracking} \kymatio is free and open-source software with a 3-clause BSD license. The members of its core development team all have experience implementing scattering transforms in other packages. A key goal of \kymatio is to combine these efforts and foster a community effort in order to produce high-quality software and maintain a critical mass of contributors for its maintenance. The package was released publicly on GitHub November 17th, 2018. The main communication channel is the GitHub page for questions, bug reports, and feature requests. There is also a dedicated Slack channel.

\begin{table}[t]
    \centering
    \begin{tabular}{l|c|c|c|c|c|c}
    & \textsc{dimension} & \textsc{gpu} & \textsc{diff.} & \textsc{core devs.} & \textsc{license} & \textsc{language} \\\hline
    ScatNet & 1D, 2D & & & 5 & Apache 2.0 & MATLAB \\
    ScatNetLight & 2D & & & 2 & GPLv2 & MATLAB \\
    PyScatWave & 2D & \checkmark & & 3 & BSD-3 & Python \\
    Scattering.m & 1D & & & 1 & GPLv3 & MATLAB \\
    PyScatHarm & 3D & \checkmark & & 1 & BSD-3 & Python \\
    Wavelet Toolbox & 1D & & & N/A & Proprietary & MATLAB \\
    \kymatio & 1D, 2D, 3D & \checkmark & \checkmark & 15 & BSD-3 & Python
    \end{tabular}
    \caption{Comparison to existing scattering transform packages.}
    \label{tab:comparison}
\end{table}

\paragraph{Relation to previous software} 
Aside from the emphasis on code quality and usability, \kymatio provides several improvements over previous scattering implementations:  
\begin{itemize}
    \item \textit{Python} is the \textit{de facto} standard for data science software, but most existing scattering packages are implemented in MATLAB. In contrast, \kymatio provides a completely Pythonic implementation, enabling integration with the scientific Python ecosystem.
    \item \textit{GPU compatibility} is critical to many data science workloads. \kymatio offers an easy-to-use GPU implementation for scattering transforms in 1D, 2D, and 3D.
    \item \textit{Frontends} are provided for many frameworks, including \textit{NumPy}, \textit{scikit-learn}, \textit{PyTorch}, and \textit{TensorFlow}/\textit{Keras}, allowing for seamless integrating scattering transforms in a variety of pipelines. In particular, the \textit{PyTorch}, and \textit{TensorFlow}/\textit{Keras} frontends allow for inclusion into many deep learning workflows.
    \item \textit{Differentiability} of the scattering transform simplifies applications in reconstruction and generative modeling, among others.
\end{itemize}

Table~\ref{tab:comparison} provides a detailed comparison of existing implementations: ScatNet \citep{scatnet}, ScatNetLight \citep{scatnetlight}, PyScatWave \citep{oyallon2018scattering}, Scattering.m \citep{lostanlen2016wavelet}, PyScatHarm \citep{eickenberg2018solid}, and the scattering transform implemented in the MATLAB Wavelet Toolbox.

\section{User Interface and Documentation}

\paragraph{Interface} The interface is designed to be flexible and consistent across inputs and frontends. Let us consider the \textit{PyTorch} frontend. We first create a scattering object by specifying the averaging scale $J$ and the input signal shape.
\begin{lstlisting}
from kymatio.torch import Scattering1D, Scattering2D, HarmonicScattering3D

S = Scattering1D(J, shape=(length,))
S = Scattering2D(J, shape=(height, width))
S = HarmonicScattering3D(J, shape=(height, width, depth))
\end{lstlisting}
The resulting object \texttt{S} acts like a \texttt{nn.Module} object in \textit{PyTorch}. The scattering transform \texttt{S} is applied through calls of the form
\begin{lstlisting}
x = torch.randn((28, 28))
output = S(x)
\end{lstlisting}
Switching from GPU or CPU functionality also follows the API of \texttt{nn.module}.
\begin{lstlisting}
S.cuda()   # Run on GPU
S.cpu()    # Run on CPU
\end{lstlisting}
\vspace{-12pt}

\paragraph{Documentation and examples}
Several examples are provided with the code, illustrating the power of \kymatio. These include image reconstruction and generation from scattering \citep{tomas}, hybrid scattering and CNN training on CIFAR and MNIST \citep{oyallon2018scattering}, regression of molecular properties on QM7/QM9 using solid harmonic scattering \citep{solidharmonic}, and classifying recordings of spoken digits.
\vspace{-10pt}
\section{Conclusion}
\kymatio provides a well documented, user-friendly, and fast implementation for the scattering transform. It can be used with the \textit{PyTorch} and \textit{TensorFlow}/\textit{Keras} deep learning frameworks and supports a variety of applications that have been previously inaccessible to non-experts including hybrid deep learning, generative modeling, and 3D chemistry applications. Future work includes further optimization for speed, flexibility, and backend support.
\vspace{-12pt}
\acks{We thank Laurent Sifre, Sergey Zagoruyko and Gabriel Huang for their helpful comments. The project was supported by ERC InvariantClass 320959. EB is funded by a Google Focused Research Award and IVADO. MJH is partially supported by Alfred P. Sloan Fellowship \#FG-2016-6607, DARPA Young Faculty Award \#D16AP00117, and NSF grant \#1620216. The Flatiron Institute is a division of the Simons Foundation.}


\vskip 0.2in

\bibliography{sample}

\end{document}